\documentclass[conference,compsoc]{IEEEtran}
\ifCLASSOPTIONcompsoc
  \usepackage[nocompress]{cite}
\else
  \usepackage{cite}
\fi
\ifCLASSINFOpdf
\else
\fi
\usepackage{graphicx}
\usepackage{booktabs}
\usepackage{multirow}
\usepackage[table ]{ xcolor}
\usepackage{subfigure}

\usepackage{pifont}

\usepackage{amsmath}
\usepackage{bbm}

\usepackage{algorithm}
\usepackage{algorithmic}

\IEEEoverridecommandlockouts

\begin{document}
%
\title{SA-Attack: Improving Adversarial Transferability of \\ Vision-Language Pre-training Models via Self-Augmentation}

\author{\IEEEauthorblockN{Bangyan He \textsuperscript{\rm 1, $^{\bullet}$ \thanks{$^{\bullet}$ First author: hebangyan@iie.ac.cn}},
Xiaojun Jia \textsuperscript{\rm 2},
Siyuan Liang \textsuperscript{\rm 3}, 
Tianrui Lou \textsuperscript{\rm 4},
Yang Liu \textsuperscript{\rm 2} and
Xiaochun Cao \textsuperscript{\rm 4, $^{\star}$ \thanks{$^{\star}$ Corresponding author: caoxiaochun@mail.sysu.edu.cn}}
}

\IEEEauthorblockA{\textsuperscript{\rm 1} Institute of Information Engineering, Chinese Academy of Sciences}

\IEEEauthorblockA{\textsuperscript{\rm 2} Nanyang Technological University}

\IEEEauthorblockA{\textsuperscript{\rm 3} National University of Singapore}

\IEEEauthorblockA{\textsuperscript{\rm 4} Sun Yat-sen University}
}

 
    


\maketitle

\begin{abstract}
Current Visual-Language Pre-training (VLP) models are vulnerable to adversarial examples. These adversarial examples present substantial security risks to VLP models, as they can leverage inherent weaknesses in the models, resulting in incorrect predictions. In contrast to white-box adversarial attacks, transfer attacks (where the adversary crafts adversarial examples on a white-box model to fool another black-box model) are more reflective of real-world scenarios, thus making them more meaningful for research. By summarizing and analyzing existing research, we identified two factors that can influence the efficacy of transfer attacks on VLP models: inter-modal interaction and data diversity. Based on these insights, we propose a self-augment-based transfer attack method, termed SA-Attack. Specifically, during the generation of adversarial images and adversarial texts, we apply different data augmentation methods to the image modality and text modality, respectively, with the aim of improving the adversarial transferability of the generated adversarial images and texts. Experiments conducted on the FLickr30K and COCO datasets have validated the effectiveness of our method. Our code will be available after this paper is accepted.
\end{abstract}


%
\IEEEpeerreviewmaketitle

\section{Introduction}

\begin{figure}[!t]
\centering
\includegraphics[width=\linewidth]{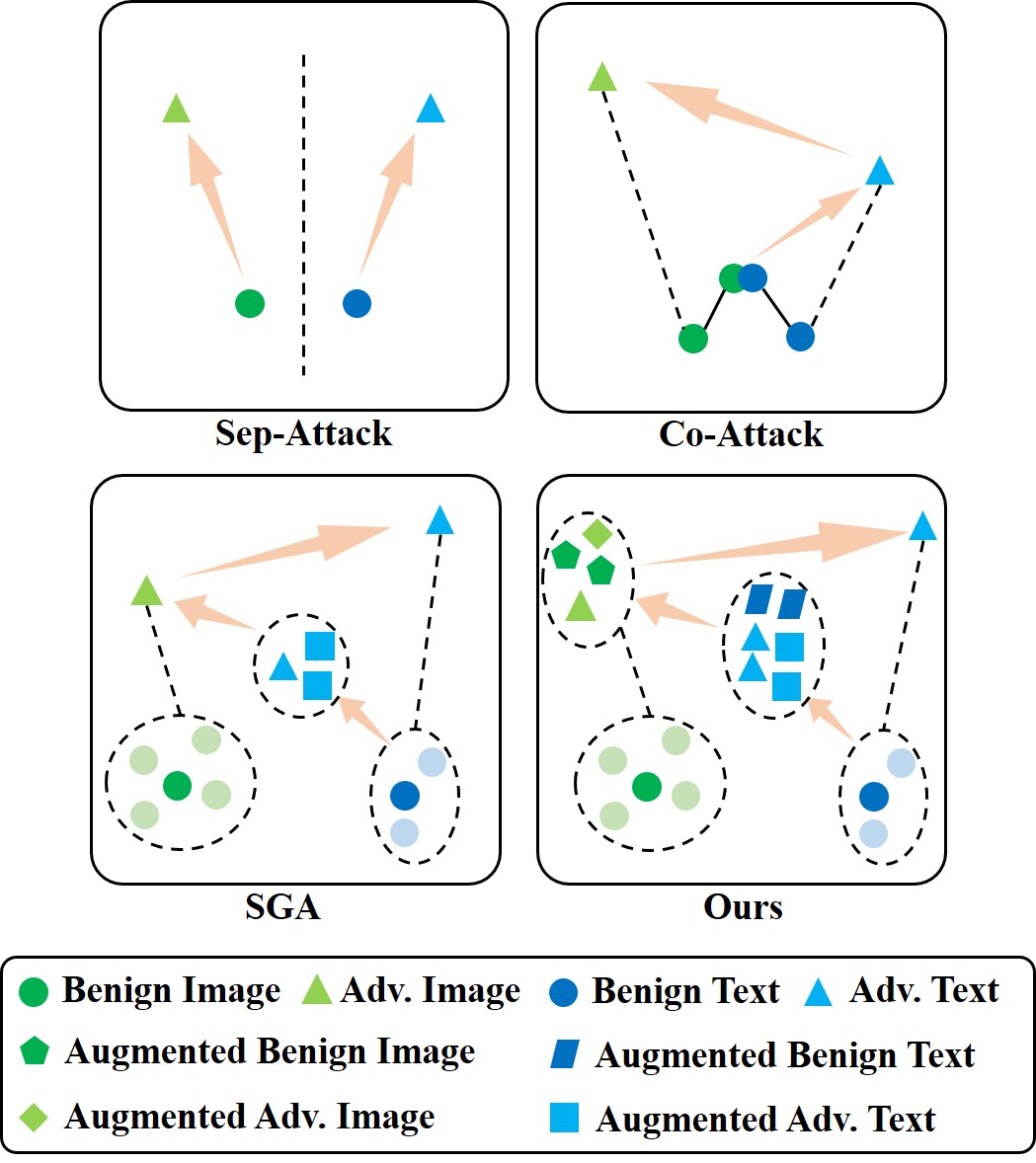}
\caption{Comparison of existing attack methods. Dots denote benign samples, while triangles denote adversarial samples. Other shapes are used to illustrate samples after different data augmentation strategies.}
\label{fig_com}
\end{figure}

Existing researches \cite{yin2023vlattack,zhou2023advclip,zhang2022towards} have shown that vision-language pre-training (VLP) models are vulnerable to adversarial examples  \cite{cao2023stylefool,hingun2023reap,jiang2023efficient}. Due to the impressive performance of the VLP models in various multi-modal tasks \cite{bao2022vlmo,wang2023position,wu2023revisiting}, their safety issues should not be underestimated.

In simple terms, an adversarial attack refers to a situation where an adversary applies imperceptible perturbations to data, causing Deep Neural Networks (DNNs) to make incorrect predictions. The data that has been perturbed by the adversary is referred to as an adversarial example. In the image modality, adversarial examples can be created by adding minimal noise \cite{xue2023diffusion} or making semantically meaningful modifications \cite{qiu2020semanticadv} to the image. In the text modality, adversarial examples can be created by altering a small portion of the words \cite{morris220textattack,li2020bert,jin2020bert}. An adversarial attack where the adversary knows and exploits the information of a DNN is referred to as a white-box attack. Conversely, a black-box attack occurs when the adversary launches an adversarial attack without knowledge of the DNN's information. In the realm of black-box attack, the topic of transfer attack has received considerable attention \cite{wei2023towards,chen2023gcma,zhang2023transferable}. A transfer attack involves an adversary generating adversarial examples on a DNN with known information, and then using these examples to attack another DNN with no known information. Compared to the white-box attack, the transfer attack is more reflective of real-world scenarios and pose a greater threat, making them a more meaningful area of study. In this paper, we primarily investigate the transfer attack on the VLP models.

Based on the existing excellent methods, i.e., Sep-Attack \cite{zhang2022towards}, Co-Attack \cite{zhang2022towards}, and SGA \cite{lu2023set}, we have summarized following two factors that can influence the efficacy of the transfer attack on VLP models: \ding{182} \textbf{Inter-modality interaction.} Zhou et al. \cite{zhou2023advclip} have pointed out that there is a gap between the image modality and the text modality. If adversarial examples are designed only for one modality, the other modality can weaken the performance of these adversarial examples. VLP models can handle both image and text modalities simultaneously, which means that the design of attack methods needs to consider the interaction between modalities. Sep-Attack did not consider the interaction between modalities, while Co-Attack did, hence its transfer attack performance is better than Sep-Attack; \ding{183} \textbf{Data diversity.} Zhu et al.\cite{zhu2022toward} have pointed out that, similar to improving the generalization of DNNs, data augmentation has a positive impact on improving adversarial transferability. Many existing works \cite{wang2023structure,byun2022improving,wang2021admix} have also confirmed the correctness of this assertion. SGA, based on Co-Attack, considered the scale-invariant characteristics of the image modality, therefore, the transfer attack performance of SGA is better than Co-Attack. However, SGA has limited consideration for data diversity, overlooking the diversity of text modality and other image attributes (e.g., structural-invariance). Based on this, we believe that taking into account a greater diversity of data could further improve the transfer attack performance on VLP models.

Motivated by the analysis mentioned above, we propose a self-augment attack method, referred to as the SA-Attack, to improve the transfer attack performance on VLP models. Here, ``self-augment'' is the term we use to denote ``augmenting the diversity of input samples''. Specifically, our method, which draws on the research of Lu et al. \cite{lu2023set}, employs a three-step process to generate transferable adversarial images and adversarial text. Initially, we input benign images and benign text into the text attack module to produce adversarial intermediate text. Subsequently, we utilize the EDA method \cite{wei2019eda} to augment the adversarial intermediate text and benign text, and then input the enhanced text into the image attack module to generate adversarial images. Finally, we apply the SIA method \cite{wang2023structure} to augment the adversarial images and benign images, and then re-input the enhanced adversarial images, benign images, and adversarial intermediate text into the text attack module to generate the final adversarial text. The image attack module is based on PGD attack \cite{madry2017towards}, while the text attack module is based on BERT-Attack \cite{li2020bert}. Figure \ref{fig_com} vividly illustrates the differences and connections between our method and previous methods in a graphical format.

To validate the effectiveness of our method, we conduct experiments using two well-established datasets, i.e., Flickr30K \cite{plummer2015flickr30k} and COCO \cite{lin2014microsoft}, and evaluate the following popular VLP models: TCL \cite{yang2022vision}, ALBEF \cite{li2021align}, CLIP$_{\text{CNN}}$ \cite{radford2021learning}, and CLIP$_{\text{ViT}}$ \cite{radford2021learning}. In addition, we use the following common transfer attack methods as baselines: PGD \cite{madry2017towards}, BERT-Attack \cite{li2020bert}, Sep-Attack \cite{zhang2022towards}, Co-Attack \cite{zhang2022towards}, and SGA \cite{lu2023set}. In our experiments, we report the attack success rates (ASRs) on R@1, R@5, and R@10, respectively. The experimental results show that our method outperforms the baselines in terms of ASRs on R@1, R@5, and R@10, respectively. This validates the effectiveness of our method across various datasets and model architectures. Furthermore, we conduct cross-task experiments, revealing the potential threat of our method in cross-task adversarial attacks. 

In summary, our main contributions are as follows.
\begin{itemize}
\item[$\bullet$] Based on previous researches, we summarize and analyze two factors that can affect the transfer attack performance on VLP models, i.e., inter-modality interaction and data diversity. 
\item[$\bullet$] Based on our analysis, we propose a self-augment attack method to improve the transfer attack performance on VLP models.
\item[$\bullet$] Experiments that we conducted on the benchmark datasets verify our attack effectiveness.
\end{itemize}
\section{Related Work}

In this section, we introduce some research work related to this paper, including the VLP model, transfer adversarial attacks, and data augmentation.

\subsection{Vision-Language Pre-training Models}
\begin{figure}[!t]
\centering
\subfigure[Single-stream]{\includegraphics[width=0.45\linewidth]{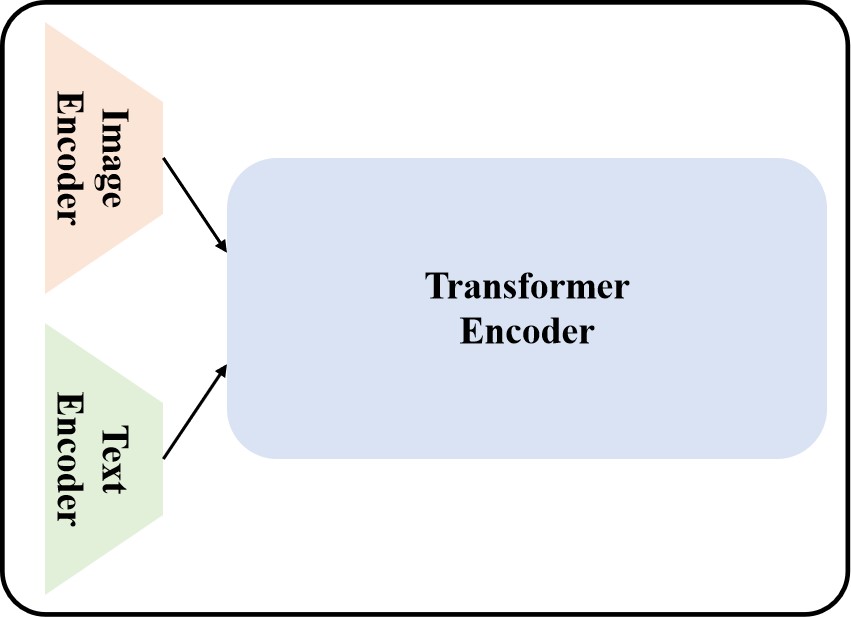}%
\label{fig_vlp_single}}
\quad
\subfigure[Dual-stream]{\includegraphics[width=0.45\linewidth]{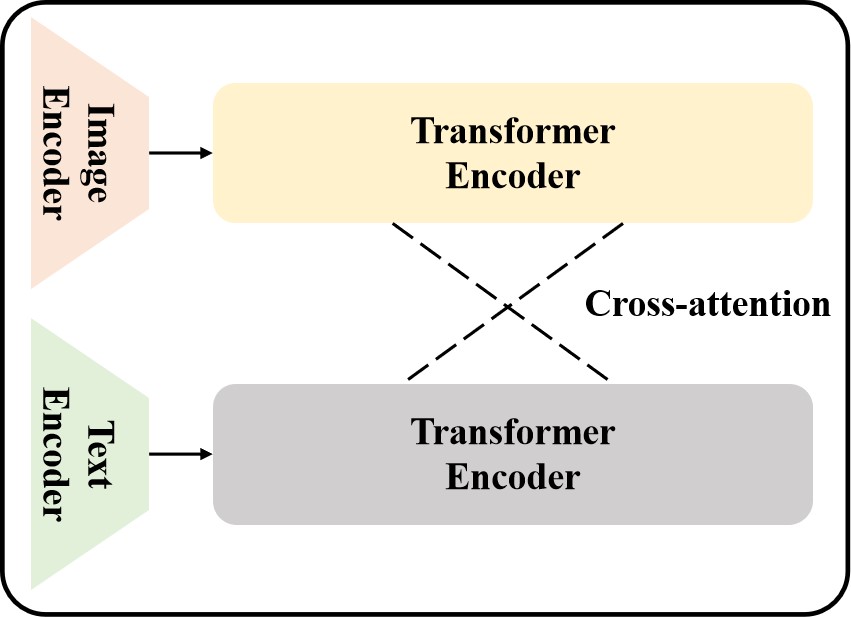}%
\label{fig_vlp_dual}}
\caption{A brief illustration of the VLP model structures. The VLP model architectures shown in Figure \ref{fig_vlp_single} concatenates textual and visual features into a shared transformer block for parameter efficiency, while the VLP model architectures shown Figure \ref{fig_vlp_dual} transmits these features to separate transformer blocks, utilizing cross-attention for enhanced performance. Different colors represent different model structures.}
\label{fig_sim}
\end{figure}

Vision-language pre-training (VLP) models leverage large-scale data to learn the semantic correspondences between different modalities \cite{chen2023vlp}, which enhances the performance of various downstream multi-modal tasks. To achieve this goal, the structure of the VLP model needs to be meticulously designed. Following \cite{chen2023vlp}, we categorize the VLP models into two types: single-stream architecture (as shown in Figure \ref{fig_vlp_single}) and dual-stream architecture (as shown in Figure \ref{fig_vlp_dual}). The single-stream architecture \cite{yang2022vision,huang2021seeing,ji2023seeing} concatenates textual and visual features, which are then input into the same transformer block. As both textual and visual modalities use the same set of parameters, the single-stream structure is more parameter-efficient than dual-stream architecture. The dual-stream architecture \cite{li2021align,radford2021learning,ji2023map} refers to a design where text features and visual features are independently transmitted to two separate transformer blocks, which do not share parameters. To enhance performance, cross-attention is employed to facilitate interaction between visual and textual modalities.

\subsection{Transferable Adversarial Attacks}
Transferable adversarial attacks refer to the adversarial examples generated on a white-box model (referred to as the source model) that can successfully deceive a black-box model (referred to as the target model). Due to the black-box nature of transferable adversarial attacks, they are more feasible in real-world scenarios and have more research significance than white-box adversarial attacks. Most transferable adversarial attacks are designed based on iterative white-box attack (i.e., I-FGSM \cite{kurakin2016adversarial} and PGD \cite{madry2017towards}). 

Following \cite{gu2023survey} and \cite{yu2023reliable}, we mainly focus on source model-based transferable adversarial attacks in this paper, and summarize existing source model-based transferable adversarial attacks into four categories: data augmentation-based, optimization-based, model-based, and loss-based.

\noindent \textbf{Data Augmentation-based Methods} use an input transformation with a stochastic component to improve the transferability of adversarial examples. Xie et al. \cite{xie2019improving} improve the transferability of adversarial examples by randomly resizing and padding of input images with a certain probability. Dong et al. \cite{dong2019evading} observe that different models use different discriminative image regions to make predictions. Based on this, they proposed a translation-invariant attack method to enhance the transferability of adversarial examples. Subsequently, Zou et al. \cite{zou2020improving} improved upon this method. Lin et al. \cite{lin2020nesterov} find that in addition to translation invariance \cite{dong2019evading}, deep neural networks also have scale-invariant property, and they propose a scale-invariant attack method. Inspired by mixup operation, Wang et al.\cite{wang2021admix} admix a small portion of each add-in images with the original image to craft more transferable adversarial examples. Byun et al. \cite{byun2022improving} draw transferable adversarial examples on 3D object. Huang et al. \cite{huang2022transferable} craft highly transferable adversarial examples via computing integrated gradients \cite{sundararajan2017axiomatic} between two images. Wang et al. \cite{wang2023structure} reveal that more diverse transformed images lead to better transferability. Based on this, they apply different transformations to different parts of the image locally, to enhance the diversity of the transformed images, and thus enhance the adversarial transferability.

\noindent \textbf{Optimization-based Methods} craft transferable adversarial examples by optimizing the process of their generation. Dong et al. \cite{dong2018boosting} Dong et al. integrate the momentum term into the iterative process of I-FGSM to improve adversarial transferability, while Lin et al. \cite{lin2020nesterov} integrate Nesterov Accelerated Gradient \cite{nesterov1983method} into I-FGSM. Zou et al. \cite{zou2022making} adapt Adam \cite{kingma20156adam} to the generation process of adversarial examples to improve their indistinguishability and transferability. Wang et al. \cite{wang2021enhancing} propose variance tuning technology that adjusts the current gradient by considering the gradient variance of the previous iteration. Qin et al. \cite{qin2022boosting} formulate the process of generating adversarial examples as a bi-level optimization problem. Zhu et al. \cite{zhu2023boosting} propose two gradient relevance frameworks to enhance adversarial transferability. 

\noindent \textbf{Model-based Methods} focus on the impact of model components on adversarial transferability. Zhou et al. \cite{zhou2018transferable} first reveal the role of intermediate features in enhancing the adversarial transferability, and many subsequent studies \cite{waseda2023closer,li2023improving,zhang2022improving} are inspired by this work. Wu et al. \cite{wu2020skip} reveal that using more gradients from skip connections based on the decay factor can craft highly transferable adversarial examples. Benz et al. \cite{benz2021batch} provide empirical evidence to support that batch normalization operations make deep neural networks more reliant on non-robust features. Naseer et al. \cite{naseer2022on} propose two strategies, Self-Ensemble and Token Refinement, to improve the adversarial transferability of existing attacks by exploiting the compositional nature of vision transformer. Huang et al. \cite{huang2023t} propose a series of self-ensemble methods for input data, attack models, and adversarial patches, aiming to improve the adversarial transferability for object detection tasks.

\noindent \textbf{Loss-based Methods} change the loss function to improve adversarial transferability, which includes but is not limited to modifying the form of the loss function and adding regularization terms. Li et al. \cite{li2020towards} introduce the Poincare distance. Xiao et al. \cite{xiao2021improving} propose a method to generate transferable adversarial patches to attack face recognition models by regularizing those adversarial patches on low-dimensional manifolds. Zhang et al. \cite{zhang2022investigating} propose normalized CE loss, which maximizing the strength of adversarial examples. Fang et al. \cite{fang2022learning} integrate meta-learning techniques in \cite{wu2020skip} to further improve the adversarial transferability. Li et al. \cite{li2023making} advocate attacking Bayesian models to achieve desirable adversarial transferability.

Currently, transferable adversarial attacks have been widely studied in various domains, including vision transformer \cite{ma2023transferable,wei2022towards,mahmood2021robustness} and face recognition \cite{li2023sibling,jia2022adv,yin2021adv}. However, transferable adversarial attacks on VLP models \cite{lu2023set,wang2023exploring} are still under development. With the growing awareness of security concerns associated with deep neural networks \cite{zhu2023ai,dyrmishi2023empirical,zhang2022adversarial}, particularly the security challenges posed by large language models \cite{liu2023prompt,deng2023jailbreaker,si2022so}, investigating the security issues of VLP models is a non-trivial task.


\subsection{Data Augmentation}
Data augmentation, a common practice in computer vision, modifies training data through a series of transformations like flipping, rotation, and cropping. This increases the training set's diversity, improving the model's generalization ability. In the image modality, MixUp \cite{zhang2017mixup} implements a linear interpolation between two distinct images, along with their corresponding labels. CutMix \cite{yun2019cutmix} involves randomly cropping a patch from another image and pasting it onto the original image. AutoAugment \cite{cubuk2018autoaugment} employs automated machine learning techniques to search for data augmentation strategies, which has inspired a series of subsequent works \cite{ho2019population,lingchen2020uniformaugment,muller2021trivialaugment}. \textbf{Structure Invariant Transformation (SIT)} \cite{wang2023structure} splits an image into many blocks. Each block undergoes a trans- formation, which is randomly selected from the following operations: VShift, HShift, VFlip, HFlip, Rotate, Scale, Add Noise, Resize, DCT, and Dropout. According to \cite{wang2023structure}, SIT can improve the diversity of images while ensuring that semantic information remains intact. In the text modality, \textbf{Easy Data Augmentation (EDA)} \cite{wei2019eda} augments text by implementing synonym replacement, random insertion, random swapping, and random deletion.

Zhu et al. \cite{zhu2022toward} point out that data augmentation has a positive impact on improving adversarial transferability. This inspires us to design methods from the perspective of data diversity, aiming to improve the adversarial transferability of VLP models.
\begin{figure*}[!t]
\centering
\includegraphics[width=0.95\linewidth]{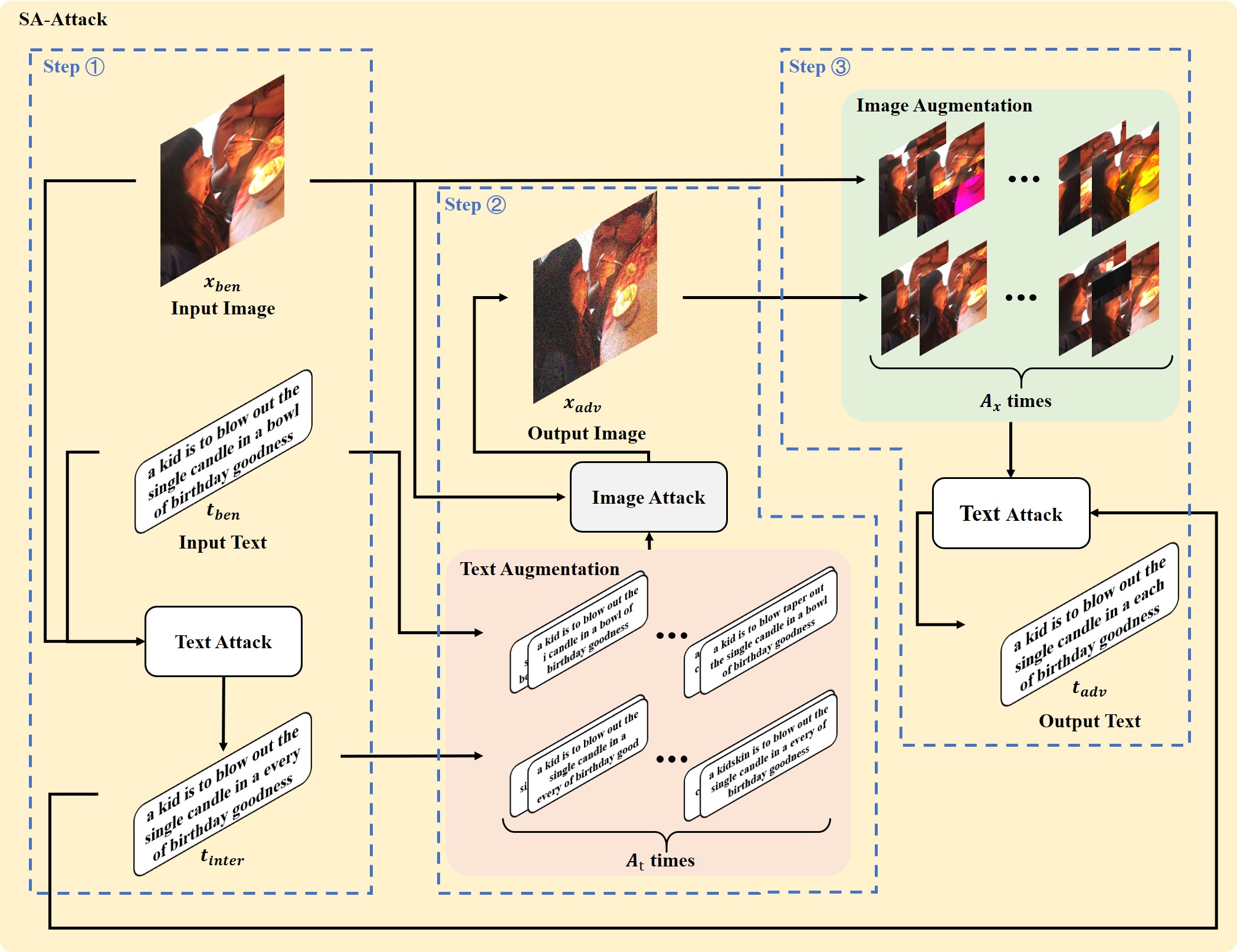}
\caption{Pipeline of our self-augment-based method, namely SA-Attack. We use the term ``self-augment'' to refer to the concept of ``augmenting the diversity of input samples''. Following \cite{lu2023set}, our method consists of three steps: \ding{182} Craft adversarial intermediate text from benign image and benign text. \ding{183} Use the augmented benign text and adversarial intermediate text, together with benign images, to craft adversarial images. \ding{184} Use the augmented benign images and adversarial images, together with adversarial intermediate text, to craft adversarial text. Different colors denote different modules. The description of each variable in the figure is shown in Table \ref{tab_notation}. Both the input image and input text are sourced from the COCO dataset \cite{lin2014microsoft}.}
\label{fig_pipeline}
\end{figure*}

\section{Methodology}

In this section, we provide a detailed description of the SA-Attack method proposed in this paper.

\subsection{Threat Model}

\noindent \textbf{Adversary’s Goals.} In this paper, we assume that the adversary generates adversarial examples in image-text retrieval. Given a white-box VLP model $f_s(\cdot)$ (referred to as the \textbf{source model}) and a benign image-text pair $(\boldsymbol{x_{ben}}, \boldsymbol{t_{ben}})$. For the image modality, the adversary generates adversarial images $\boldsymbol{x_{adv}}$ using the following equation:
\begin{equation}
\boldsymbol{x_{adv}} = \underset{\boldsymbol{x_{adv}} \in \mathcal{B}[\boldsymbol{x_{ben}}, \epsilon_{x}]}{\arg \max } \mathcal{L}_{i}\left(f_s(\boldsymbol{x_{ben}})\right),
\end{equation}
where $\mathcal{L}_{i}(\cdot)$ denotes loss function designed by the adversary for the image modality, $\mathcal{B}[\boldsymbol{x_{ben}}, \epsilon_{x}]$ denotes the legal searching spaces for optimizing adversarial image, and $\epsilon_{x}$ denotes the maximum range that $\boldsymbol{x_{adv}}$ can be modified.


For the text modality, the adversary generates adversarial text $\boldsymbol{t_{adv}}$ using the the following equation:
\begin{equation}
\boldsymbol{t_{adv}} = \underset{\boldsymbol{t_{adv}} \in \mathcal{B}[\boldsymbol{t_{ben}}, \epsilon_{t}]}{\arg \max } \mathcal{L}_{t}\left(f_s(\boldsymbol{t_{ben}})\right),
\end{equation}
where $\mathcal{L}_{t}(\cdot)$ denotes loss function designed by the adversary for the text modality, $\mathcal{B}[\boldsymbol{t_{ben}}, \epsilon_{t}]$ denotes the legal searching spaces for optimizing adversarial text, and $\epsilon_{t}$ denotes the maximum range that $\boldsymbol{t_{adv}}$ can be modified.

Then, given a black-box VLP model $f_t(\cdot)$ (referred to as the \textbf{target model}), the adversary's goal is to input $\boldsymbol{x_{adv}}$ into $f_t(\cdot)$ such that none of the top-k results returned by $f_t(\cdot)$ are correct. Similarly, the adversary aims to achieve the same effect with $\boldsymbol{t_{adv}}$ inputs into $f_t(\cdot)$, where none of the top-k results returned by $f_t(\cdot)$ are correct.

\noindent \textbf{Adversary’s Capabilities.} Following previous researches \cite{lu2023set,byun2022improving,wang2021admix}, we assume that both $f_s(\cdot)$ and $f_t(\cdot)$ are trained on the same dataset. The adversary has access to all information about $f_s(\cdot)$, such as model structure, parameters, and the dataset used for adversarial attacks. However, the adversary is completely unaware of $f_t(\cdot)$ and cannot obtain any information about it. This setting presents a certain level of challenge to our research and is more closely aligned with real-world scenarios.

\subsection{Motivation} \label{sec_motivation}

Based on previous researches \cite{zhou2023advclip,yin2023vlattack,ban2022pre}, we have identified two factors that may influence adversarial transferability: \textbf{inter-modality interaction and data diversity}. With these two points in mind, we discuss the limitations of existing transfer attack methods on VLP models, i.e., Sep-Attack \cite{zhang2022towards}, Co-Attack \cite{zhang2022towards}, and SGA \cite{lu2023set}. Figure \ref{fig_com} provides a comparison of these methods in a graphical format.

\noindent \textbf{Sep-Attack} overlooks the interaction between the image and text modalities. Existing research \cite{zhou2023advclip} indicates that a gap exists between the image and text modalities. To be specific, if adversarial examples can only be crafted in either the image modality or the text modality, VLP models can utilize information from the other modality, thereby reducing the attack performance of these adversarial examples. When designing attack methods for VLP models, which encompass both text and image modalities, this gap's neglect could potentially affect the adversarial transferability.

\noindent \textbf{Co-Attack} fails to consider the diversity among image-text pairs. Although Co-Attack takes into account the interaction between multi-modalities, it only uses a single image-text pair to craft adversarial examples. Many studies \cite{wang2023structure,byun2022improving,wang2021admix} have demonstrated that the diversity of input data is a crucial factor influencing adversarial transferability.

\noindent \textbf{SGA} utilizes a limited diversity among image-text pairs. In comparison to the first two attack methods, SGA not only considers the interaction between multi-modalities but also acknowledges the diversity of input data. However, SGA only contemplates scale-invariance in images. The diversity of the text modality, as well as other properties of image modalities (such as translation invariance, structural invariance), remain unconsidered.

The above analysis suggests that we should imporove adversarial transferability from the perspective of data diversity. The experiments conducted in Section \ref{sec_exp_transfer} further validate the accuracy of our analysis.

\subsection{Pipeline}
In this subsection, we first outline the pipeline of our self-augment-based method, which we have named \textbf{SA-Attack}. In this paper, The concept of ``augmenting the variety of input samples'' is what we mean by ``self-augment''. Following this, we introduce the text attack and image attack modules within SA-Attack in detail.

\noindent \textbf{Overview.} Figure \ref{fig_pipeline} shows our pipeline, and Table \ref{tab_notation} describes the variables in Figure \ref{fig_pipeline}. Similar to \cite{lu2023set}, we also adopt a three-step process to generate transferable adversarial images and adversarial text. \ding{182} Benign images \textbf{$\boldsymbol{x_{ben}}$} and benign texts \textbf{$\boldsymbol{t_{ben}}$} are input into text attack module to craft adversarial intermediate text \textbf{$\boldsymbol{t_{inter}}$}. \ding{183} \textbf{$\boldsymbol{t_{inter}}$}, along with \textbf{$\boldsymbol{t_{ben}}$}, undergo augmentation using EDA and are then input into the image attack module, resulting in adversarial images \textbf{$\boldsymbol{x_{adv}}$}. \ding{184} Adversarial images \textbf{$\boldsymbol{x_{adv}}$} and benign images \textbf{$\boldsymbol{x_{ben}}$} are augmented using SIA. Along with the adversarial intermediate texts, the augmented \textbf{$\boldsymbol{x_{adv}}$} and \textbf{$\boldsymbol{x_{ben}}$} are re-input into the text attack module to craft the final adversarial texts. 

The text attack module and the image attack module will be described in the following.

\begin{table}[!t]
\caption{Descriptions of each variable depicted in Figure \ref{fig_pipeline}.}
\label{tab_notation}
\centering
\begin{tabular}{@{}cc@{}}
\toprule
\textbf{Variable}           & \textbf{Description}            \\ \midrule
\textbf{$\boldsymbol{x_{ben}}$}          & Benign image             \\
\textbf{$\boldsymbol{x_{adv}}$}          & Adversarial image\\
\textbf{$A_{x}$}     & The number of times image is augmented \\
\textbf{$\boldsymbol{t_{ben}}$}          & Benign text               \\
\textbf{$\boldsymbol{t_{inter}}$} & Adversarial   intermediate text \\
\textbf{$\boldsymbol{t_{adv}}$}          & Adversarial text       \\
\textbf{$A_{t}$}     & The number of times text is augmented \\
\bottomrule
\end{tabular}
\end{table}

\begin{algorithm}[!t]
\renewcommand{\algorithmicrequire}{\textbf{Input:}}
\renewcommand{\algorithmicensure}{\textbf{Output:}}
\caption{Text Attack}
\label{alg_txt}
\begin{algorithmic}[1]
\REQUIRE ~~\\
Input texts with $n$ words $t_{in}=[w_0, w_1, \cdots, w_n]$; \\
Adversarial intermediate texts $\boldsymbol{t_{inter}}$; \\
Benign image $\boldsymbol{x_{ben}}$; \\
Augmented benign images \\ $x_{ben}^{A} = \{\boldsymbol{x_{ben}^{1}}, \boldsymbol{x_{ben}^{2}}, \cdots , \boldsymbol{x_{ben}^{A_{x}}}\}$; \\
Augmented adversarial images \\ $x_{adv}^{A} = \{\boldsymbol{x_{adv}^{1}}, \boldsymbol{x_{adv}^{2}}, \cdots , \boldsymbol{x_{adv}^{A_{x}}}\}$; \\
Source model $f_s$; \\
Top-$k$ value.
\ENSURE ~~\\
Adversarial text $\boldsymbol{t_{adv}}$.
\STATE Select top-$k$ important word list in $t_{in}$: \\ $L=[w_{\text{top-1}}, w_{\text{top-2}}, \cdots, w_{\text{top-k}}]$; \\ /\ Use word importance ranking from \cite{li2020bert}. /\
\IF{$x_{ben}^{A} \cup x_{adv}^{A}$ is not $None$}
    \STATE Concatenate augmented images: \\ $x_{cat} \gets \text{Concatenate}(x_{ben}^{A}, x_{adv}^{A})$;
\ENDIF
\STATE Initialize a list: $I \gets [ \quad ]$;
\FOR{each $w_i$ in $L$}
    \STATE Replace the $i$-th word of $t_{in}$: \\ $\boldsymbol{t_{i}^{\prime}} \gets \text{Replace}_{i}(t_{in})$; \\ /\ Use word replacement strategy from \cite{li2020bert}. /\
    \IF{$x_{cat}$ is not $None$}
        \STATE Calculate score $\mathcal{L}_{t}$ using Equation \ref{eq_lt_not_none};
    \ELSE
        \STATE Calculate score $\mathcal{L}_{t}$ using Equation \ref{eq_lt_none};
    \ENDIF
    \STATE $I.\text{append}(\mathcal{L}_{t})$;
\ENDFOR
\STATE Obtain the index of the maximum value $l$ in $I$;
\STATE  $\boldsymbol{t_{adv}} \gets \boldsymbol{t_{l}^{\prime}}$.
\RETURN $\boldsymbol{t_{adv}}$
\end{algorithmic}
\end{algorithm}

\begin{algorithm}[!t]
\renewcommand{\algorithmicrequire}{\textbf{Input:}}
\renewcommand{\algorithmicensure}{\textbf{Output:}}
\caption{Image Attack}
\label{alg_img}
\begin{algorithmic}[1]
\REQUIRE ~~\\
Benign image $\boldsymbol{x_{ben}}$; \\
Benign texts $\boldsymbol{t_{ben}}$; \\
Augmented benign texts \\ $t_{ben}^{A} = \{\boldsymbol{t_{ben}^{1}}, \boldsymbol{t_{ben}^{2}}, \cdots, \boldsymbol{t_{ben}^{A_{t}}}\}$; \\
Adversarial intermediate texts \\ $t_{inter}^{A} = \{\boldsymbol{t_{inter}^{1}}, \boldsymbol{t_{inter}^{2}}, \cdots, \boldsymbol{t_{inter}^{A_{t}}}\}$; \\
Source model $f_s$; \\
Maximum perturbation $\epsilon_{x}$; \\
Step size $\alpha$; \\
Number of iterations $T$.
\ENSURE Adversarial image $\boldsymbol{x_{adv}}$.
\STATE Initialize $\boldsymbol{x_{adv}} \gets \boldsymbol{x_{ben}} + \boldsymbol{\delta}$;
\STATE $\boldsymbol{x_{adv}} \gets \text{Clip}_{[0, 1]}(\boldsymbol{x_{adv}})$;
\STATE Concatenate augmented texts \\ $\boldsymbol{t_{cat}} \gets \text{Concatenate}(t_{ben}^{A}, t_{inter}^{A})$;
\FOR{$t=1$ to $T$}
\STATE Get scaled images: $\boldsymbol{x_{adv}}^{s} \gets \text{Scale}(\boldsymbol{x_{adv}})$; \\ /\ Use scale image method in \cite{lu2023set}. /\;
\STATE $\boldsymbol{g} \gets \nabla_{\boldsymbol{x_{adv}}} \mathcal{L}_{i}(\boldsymbol{x_{adv}}, \boldsymbol{t_{ben}}, \boldsymbol{t_{cat}}, f_s)$; \\ /\ The expression of $\mathcal{L}_{i}(\cdot)$ is shown in Equation \ref{eq:loss_img}. /\
\STATE $\boldsymbol{x_{adv}} \gets \boldsymbol{x_{adv}} + \alpha \cdot \text{sign}(\boldsymbol{g})$;
\STATE $\boldsymbol{x_{adv}} \gets \text{Clip}_{[\boldsymbol{x_{adv}}-\epsilon_{x}, \boldsymbol{x_{adv}}+\epsilon_{x}]}(\boldsymbol{x_{adv}})$;
\STATE $\boldsymbol{x_{adv}} \gets \text{Clip}_{[0, 1]}(\boldsymbol{x_{adv}})$.
\ENDFOR
\RETURN $\boldsymbol{x_{adv}}$
\end{algorithmic}
\end{algorithm}

\noindent \textbf{Text Attack.} Following \cite{lu2023set}, we have modified BERT-Attack \cite{li2020bert} as the basis for our text attack module. Simply put, BERT-Attack consists of two steps: \ding{182} Identifying the words in a given input text that are vulnerable to attack, and \ding{183} Using BERT to craft substitutes for these vulnerable words. Algorithm \ref{alg_txt} provides a detailed illustration of the process for the text attack module.

Firstly, we employ the word importance ranking strategy from \cite{li2020bert} to identify words that are vulnerable to attacks. These words are then sorted in descending order of importance, and the top-k important words are selected to form a list $L=\{w_{top-1}, w_{top-2}, ..., w_{top-k}\}$ (line 1). If $x_{ben}^{A}$ and $x_{adv}^{A}$ are not None, they are concatenated; if not, no action is taken (lines 2 to 4). 

Next, for each important word $w_i$ in $L$, the following steps are performed: 
\ding{182} The $i$-th word in $t_{in}$ is replaced with the word replacement strategy from \cite{li2020bert}, resulting in the modified text $\boldsymbol{t_{i}^{\prime}}$ (line 7).
\ding{183} If $x_{cat}$ is None, $\mathcal{L}_{t}$ is calculated using the following equation:
\begin{equation} \label{eq_lt_none}
\mathcal{L}_{t} = -\frac{f_{s}\left(\boldsymbol{t_{i}^{\prime}}\right) \cdot f_{s}(\boldsymbol{x_{ben}})}{\left\|f_{s}\left(\boldsymbol{t_{i}^{\prime}}\right)\right\| \cdot \left\|f_{s}(\boldsymbol{x_{ben}})\right\|},
\end{equation}
where $f_s$ denotes the source model. When a text is input into $f_s$, it outputs text features. Similarly, when an image is input into $f_s$, it outputs image features; otherwise, $\mathcal{L}_{t}$ is calculated using the following equation (lines 8 to 12):
\begin{equation} \label{eq_lt_not_none}
\begin{aligned} 
\mathcal{L}_{t} = &- \frac{f_{s}\left(\boldsymbol{t_{i}^{\prime}}\right) \cdot f_{s}\left(\boldsymbol{x_{ben}}\right)}{\left\|f_{s}\left(\boldsymbol{t_{i}^{\prime}}\right)\right\| \cdot \left\|f_{s}\left(\boldsymbol{x_{ben}}\right)\right\|} \\
& - \sum_{j=1}^{2 \cdot A_{x}}\left(\frac{f_{s}\left(\boldsymbol{t_{i}^{\prime}}\right) \cdot f_{s}\left(\boldsymbol{x_{cat}^{j}}\right)}{\left\|f_{s}\left(\boldsymbol{t_{i}^{\prime}}\right)\right\| \cdot \left\|f_{s}\left(\boldsymbol{x_{cat}^{j}}\right)\right\|}\right).
\end{aligned}
\end{equation}
The goal of Equation \ref{eq_lt_none} is to maximize the distance between $\boldsymbol{t_{i}^{\prime}}$ and $\boldsymbol{x_{ben}}$ in the high-dimensional manifold. The goal of Equation \ref{eq_lt_not_none} is to maximize the distance between $\boldsymbol{t_{i}^{\prime}}$ and both $\boldsymbol{x_{ben}}$ and $x_{cat}$ within a high-dimensional manifold. \ding{184} $\mathcal{L}_{t}$ is then added to list $I$ (line 13).

After all the words in $L$ have been processed, the index $l$ of the maximum value in $L$ is found (line 15). The adversarial text $\boldsymbol{t_{adv}}$ is then set as the text obtained by replacing the word at the corresponding index position in $\boldsymbol{t_{adv}}$ (line 16), i.e., $\boldsymbol{t_{adv}} \gets \boldsymbol{t_{l}^{\prime}}$.

\begin{table*}[!t]
\caption{Comparison with existing attack methods on image-text retrieval task. This table reports the attack success rate (\%) R@1 for IR and TR. The gray background indicates white-box attacks. The results of Co-Attack are reported in \cite{lu2023set}. A higher ASR in black-box attack indicates better adversarial transferability. The best results are indicated in bold.}
\label{tab_trans_1}
\centering
\begin{tabular}{@{}cccccccccc@{}}
\toprule
\multicolumn{10}{c}{Flickr30K Dataset}                                                                                                                                                                                                                                                                                                                                                                                                        \\ \midrule
\multicolumn{2}{c|}{Target Model ($\rightarrow$)}                                                              & \multicolumn{2}{c|}{ALBEF}                                                         & \multicolumn{2}{c|}{TCL}                                                           & \multicolumn{2}{c|}{CLIP$_{\text{ViT}}$}                                           & \multicolumn{2}{c}{CLIP$_{\text{CNN}}$}                       \\ \midrule
\multicolumn{1}{c|}{Source Model   ($\downarrow$)}         & \multicolumn{1}{c|}{Attack Method ($\downarrow$)} & TR R@1                        & \multicolumn{1}{c|}{IR R@1}                        & TR R@1                        & \multicolumn{1}{c|}{IR R@1}                        & TR R@1                        & \multicolumn{1}{c|}{IR R@1}                        & TR R@1                        & IR R@1                        \\ \midrule
\multicolumn{1}{c|}{}                                      & \multicolumn{1}{c|}{PGD}                          & \cellcolor[HTML]{EFEFEF}56.52 & \multicolumn{1}{c|}{\cellcolor[HTML]{EFEFEF}62.37} & 4.00                          & \multicolumn{1}{c|}{7.64}                          & 8.34                          & \multicolumn{1}{c|}{13.34}                         & 9.71                          & 14.96                         \\
\bottomrule
\end{tabular}
\end{table*}

\begin{table*}[!t]
\caption{Comparison with existing attack methods on image-text retrieval task. This table reports the attack success rate (\%) R@5 for IR and TR. The gray background indicates white-box attacks. The results of Co-Attack are reported in \cite{lu2023set}. A higher ASR in black-box attack indicates better adversarial transferability. The best results are indicated in bold.}
\label{tab_trans_5}
\centering
\begin{tabular}{@{}cccccccccc@{}}
\toprule
\multicolumn{10}{c}{Flickr30K Dataset}                                                                                                                                                                                                                                                                                                                                                                                                                                                                                \\ \midrule
\multicolumn{2}{c|}{Target Model ($\rightarrow$)}                                                              & \multicolumn{2}{c|}{ALBEF}                                                                           & \multicolumn{2}{c|}{TCL}                                                                             & \multicolumn{2}{c|}{CLIP$_{\text{ViT}}$}                                                             & \multicolumn{2}{c}{CLIP$_{\text{CNN}}$}                                         \\ \midrule
\multicolumn{1}{c|}{Source Model   ($\downarrow$)}         & \multicolumn{1}{c|}{Attack Method ($\downarrow$)} & TR R@5                                 & \multicolumn{1}{c|}{IR R@5}                                 & TR R@5                                 & \multicolumn{1}{c|}{IR R@5}                                 & TR R@5                                 & \multicolumn{1}{c|}{IR R@5}                                 & TR R@5                                 & IR R@5                                 \\
\bottomrule
\end{tabular}
\end{table*}

\begin{table*}[!t]
\caption{Comparison with existing attack methods on image-text retrieval task. This table reports the attack success rate (\%) R@10 for IR and TR. The gray background indicates white-box attacks. The results of Co-Attack are reported in \cite{lu2023set}. A higher ASR in black-box attack indicates better adversarial transferability. The best results are indicated in bold.}
\label{tab_trans_10}
\centering
\begin{tabular}{@{}cccccccccc@{}}
\toprule
\multicolumn{10}{c}{Flickr30K Dataset}                                                                                                                                                                                                                                                                                                                                                                                                        \\ \midrule
\multicolumn{2}{c|}{Target Model ($\rightarrow$)}                                                              & \multicolumn{2}{c|}{ALBEF}                                                         & \multicolumn{2}{c|}{TCL}                                                           & \multicolumn{2}{c|}{CLIP$_{\text{ViT}}$}                                           & \multicolumn{2}{c}{CLIP$_{\text{CNN}}$}                       \\ \midrule
\multicolumn{1}{c|}{Source Model   ($\downarrow$)}         & \multicolumn{1}{c|}{Attack Method ($\downarrow$)} & TR R@10                       & \multicolumn{1}{c|}{IR R@10}                       & TR R@10                       & \multicolumn{1}{c|}{IR R@10}                       & TR R@10                       & \multicolumn{1}{c|}{IR R@10}                       & TR R@10                       & IR R@10                       \\ \midrule
\multicolumn{1}{c|}{}                                      & \multicolumn{1}{c|}{PGD}                          & \cellcolor[HTML]{E7E6E6}33.00 & \multicolumn{1}{c|}{\cellcolor[HTML]{E7E6E6}40.88} & \cellcolor[HTML]{FFFFFF}0.20  & \multicolumn{1}{c|}{\cellcolor[HTML]{FFFFFF}1.46}  & 0.20                          & \multicolumn{1}{c|}{3.14}                          & 1.85                          & 3.50                          \\
\bottomrule
\end{tabular}
\end{table*}

\begin{table*}[!t]
\caption{Ablation Studies on $A_x$. This table reports the attack success rate (\%) R@1 for IR and TR. The value of $A_t$ is fixed at $4$. The gray background indicates white-box attacks. The best results are indicated in bold.}
\label{tab_ab_ax}
\centering
\begin{tabular}{@{}cccccccccc@{}}
\toprule
\multicolumn{10}{c}{Flickr30K Dataset}                                                                                                                                                                                                                                                                                                                                                                                                                                    \\ \midrule
\multicolumn{2}{c|}{Target Model ($\rightarrow$)}                                                      & \multicolumn{2}{c|}{ALBEF}                                                                           & \multicolumn{2}{c|}{TCL}                                                                             & \multicolumn{2}{c|}{CLIP$_{\text{ViT}}$}                                           & \multicolumn{2}{c}{CLIP$_{\text{CNN}}$}                       \\ \midrule
\multicolumn{1}{c|}{Source Model ($\downarrow$)}           & \multicolumn{1}{c|}{$A_x$ ($\downarrow$)} & TR R@1                                 & \multicolumn{1}{c|}{IR R@1}                                 & TR R@1                                 & \multicolumn{1}{c|}{IR R@1}                                 & TR R@1                        & \multicolumn{1}{c|}{IR R@1}                        & TR R@1                        & IR R@1                        \\ \midrule
\multicolumn{1}{c|}{}                                      & \multicolumn{1}{c|}{4}                    & \cellcolor[HTML]{E7E6E6}96.56          & \multicolumn{1}{c|}{\cellcolor[HTML]{E7E6E6}96.54}          & \textbf{48.16}                         & \multicolumn{1}{c|}{57.90}                                  & \textbf{35.34}                & \multicolumn{1}{c|}{47.13}                         & 37.93                         & 49.33                         \\
\bottomrule
\end{tabular}
\end{table*}

\begin{table*}[!t]
\caption{Ablation Studies on $A_t$. This table reports the attack success rate (\%) R@1 for IR and TR. The value of $A_x$ is fixed at $4$. The gray background indicates white-box attacks. The best results are indicated in bold.}
\label{tab_ab_at}
\centering
\begin{tabular}{@{}cccccccccc@{}}
\toprule
\multicolumn{10}{c}{Flickr30K Dataset}                                                                                                                                                                                                                                                                                                                                                                                                \\ \midrule
\multicolumn{2}{c|}{Target Model ($\rightarrow$)}                                                      & \multicolumn{2}{c|}{ALBEF}                                                         & \multicolumn{2}{c|}{TCL}                                                           & \multicolumn{2}{c|}{CLIP$_{\text{ViT}}$}                                           & \multicolumn{2}{c}{CLIP$_{\text{CNN}}$}                       \\ \midrule
\multicolumn{1}{c|}{Source Model ($\downarrow$)}           & \multicolumn{1}{c|}{$A_t$ ($\downarrow$)} & TR R@1                        & \multicolumn{1}{c|}{IR R@1}                        & TR R@1                        & \multicolumn{1}{c|}{IR R@1}                        & TR R@1                        & \multicolumn{1}{c|}{IR R@1}                        & TR R@1                        & IR R@1                        \\ \midrule
\multicolumn{1}{c|}{}                                      & \multicolumn{1}{c|}{4}                    & \cellcolor[HTML]{EFEFEF}96.56 & \multicolumn{1}{c|}{\cellcolor[HTML]{EFEFEF}96.54} & \textbf{48.16}                & \multicolumn{1}{c|}{\textbf{57.90}}                & \textbf{35.34}                & \multicolumn{1}{c|}{\textbf{47.13}}                & \textbf{37.93}                & \textbf{49.33}                \\
\bottomrule
\end{tabular}
\end{table*}

\begin{table}[!t]
\caption{Cross-Task Transferability: ITR $\rightarrow$ VG. A lower value signifies superior adversarial transferability.}
\label{tab_vg}
\centering
\begin{tabular}{@{}c|ccc@{}}
\toprule
Attack   Mtehod ($\downarrow$) & Val    & Test A & Test B \\ \midrule
Baseline                       & 0.5689 & 0.6437 & 0.4584 \\ 
SA-Attack   (Ours)             & \textbf{0.5621} & \textbf{0.6338} & \textbf{0.4524} \\ \bottomrule
\end{tabular}
\end{table}

\begin{table}[!t]
\caption{Time Comparison. The dataset used is Flickr30K, the source model is ALBEF, and the target model is TCL.}
\label{tab_time}
\centering
\begin{tabular}{@{}c|ccc@{}}
\toprule
$A_t (A_x=4)$      & 4    & 8    & 12   \\ \midrule
Time Cost (second) & 2520 & 2572 & 2692 \\ \hline\hline
$A_x (A_t=4)$      & 4    & 8    & 12   \\ \midrule
Time Cost (second) & 2520 & 2534 & 2584 \\ \bottomrule
\end{tabular}
\end{table}

\begin{figure*}[!ht]
\centering
\includegraphics[width=0.97\linewidth]{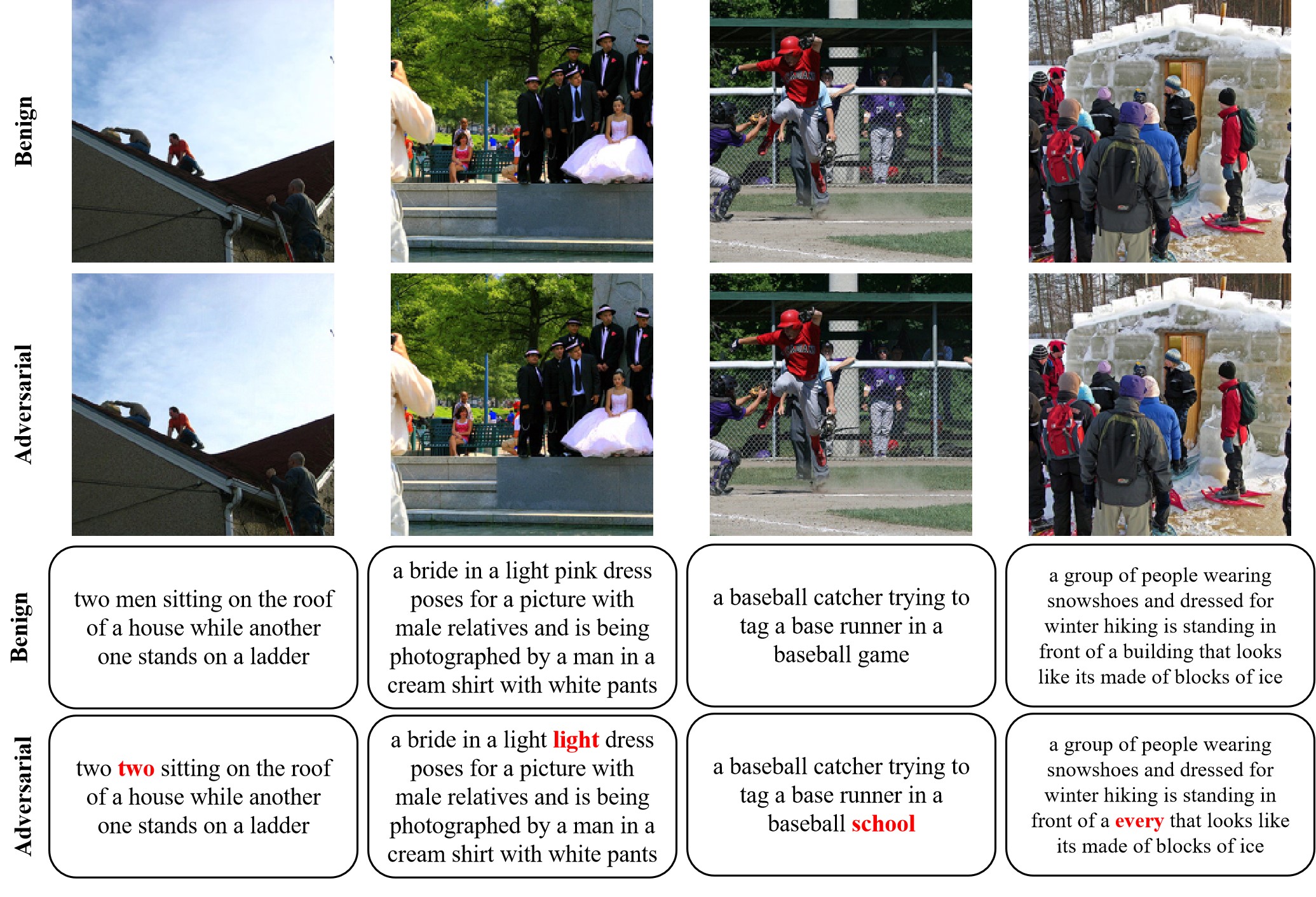}
\caption{Visualization of our method on the Flickr30K dataset. In the listed adversarial text, the modified words are displayed in red bold font.}
\label{fig_vis_flickr}
\end{figure*}

\begin{figure*}[!ht]
\centering
\includegraphics[width=0.97\linewidth]{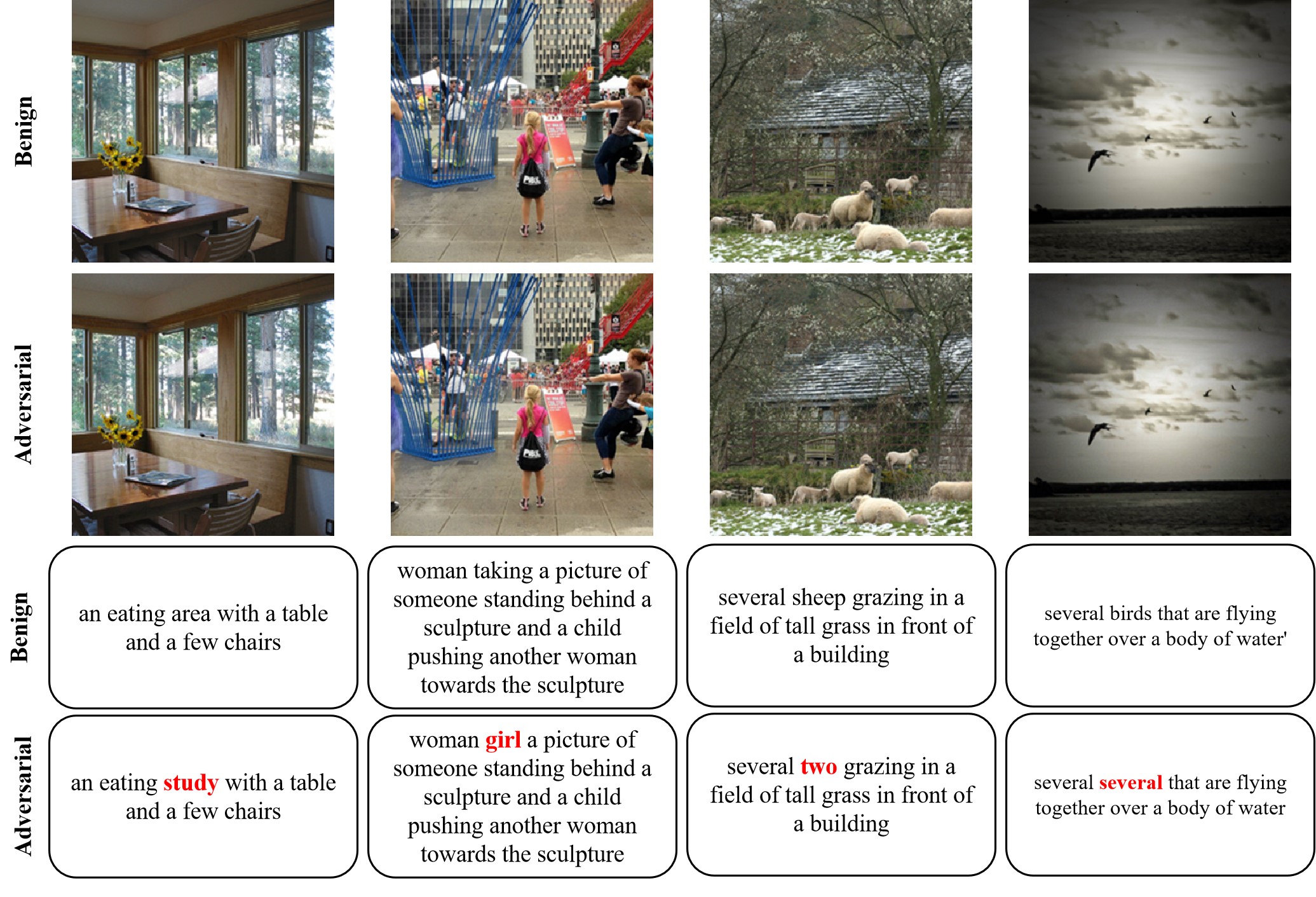}
\caption{Visualization of our method on the COCO dataset. In the listed adversarial text, the modified words are displayed in red bold font.}
\label{fig_vis_coco}
\end{figure*}

\noindent \textbf{Image Attack.} Algorithm \ref{alg_img} shows the detailed process of the image attack module. Specifically, we employ the PGD-like attack method to generate adversarial images. 

Prior to the start of the iteration (lines 1 to 3), we initialize the adversarial samples by adding a uniform distribution with a mean of 0 and a variance of 0.05 to the benign image $\boldsymbol{x_{ben}}$ (line 1), i.e.,
\begin{equation}
\boldsymbol{x_{adv}} \gets \boldsymbol{x_{ben}} + \boldsymbol{\delta}, \quad \boldsymbol{\delta} \sim U(0, 0.05).
\end{equation}
Then, we confine the initialized $\boldsymbol{x_{adv}}$ within the range of $[0,1]$ (line2), i.e.,
\begin{equation}
\boldsymbol{x_{adv}} \gets \text{Clip}_{[0, 1]}(\boldsymbol{x_{adv}}),
\end{equation}
where $\text{Clip}_{[a, b]}(\cdot)$ denotes an operation that limits the minimum and maximum values of an input tensor to $a$ and $b$, respectively. Subsequently, we concatenate augmented benign texts $\boldsymbol{t_{ben}}^{i} (i = 1,2,\cdots, A_{t})$ and adversarial intermediate texts $\boldsymbol{t_{inter}}^{i} (i = 1,2,\cdots, A_{t})$ to obtain $\boldsymbol{t_{cat}}$ (line3).

During the iteration process (lines 4 to 9), we first utilize the scale operation in \cite{lu2023set} to obtain the scaled image $\boldsymbol{x_{adv}}^{s}$ (line 5). Then, we use the following loss function to obtain the gradient $g$ (line 6).
\begin{equation} \label{eq:loss_img}
\begin{aligned}
\mathcal{L}_{i} = &- \frac{f_{s}\left(\boldsymbol{t_{ben}}\right)}{\left\|f_{s}\left(\boldsymbol{t_{ben}}\right)\right\|} \sum_{\boldsymbol{x_{i}}  \subseteq \boldsymbol{x_{adv}}^{s}} \frac{f_{s}\left(\boldsymbol{x_i}\right)}{\left\|f_{s}\left(\boldsymbol{x_i}\right)\right\|} \\
&-\sum_{j=1}^{2 \cdot A_{t}} \left( \frac{f_{s}\left(\boldsymbol{t^{j}_{cat}}\right)}{\left\|f_{s}\left(\boldsymbol{t^{j}_{cat}}\right)\right\|} \cdot  \sum_{\boldsymbol{x_{i}}  \subseteq \boldsymbol{x_{adv}}^{s}} \frac{f_{s}\left(\boldsymbol{x_i}\right)}{\left\|f_{s}\left(\boldsymbol{x_i}\right)\right\|}\right ).
\end{aligned}
\end{equation}
The goal of Equation \ref{eq:loss_img} is to encourage that the adversarial images $\boldsymbol{x_{adv}}$ are as far away as possible from $\boldsymbol{t_{ben}}$, $t_{ben}^{A}$ and $t_{inter}^{A}$ in the high-dimensional manifold. Finally, we employ the gradient ascent method in PGD to obtain the perturbations (line 7), and we constrain the magnitude of the perturbations within $\epsilon_{x}$ using the $\ell_{\infty}$ norm (line 8).


\section{Experiments}

In this section, we conduct experiments to validate the effectiveness of our method, where the datasets used in the experiments, models, baselines, hyper-parameter settings, and experimental results are elaborated in detail.

\subsection{Setup} \label{sec_setup}

\noindent \textbf{Datasets.} We conduct our experiments on two well-established datasets, i.e., Flickr30K \cite{plummer2015flickr30k} and COCO \cite{lin2014microsoft}. The Flickr30K dataset contains 31,000 images, each accompanied by five reference sentences. This rich combination of visual and linguistic information provides a valuable resource for researchers, and is widely used in fields such as image description and cross-modal retrieval. The first version of the COCO dataset, released in 2014, includes 164K images divided into training set (83K images), validation set (41K images), and testing set (41K images). In addition, the COCO dataset provides extensive annotation information, making it widely applicable in many tasks such as object detection, instance segmentation, and image captioning.

\noindent \textbf{Models.} Following \cite{lu2023set}, we evaluate three popular VLP models, i.e., TCL \cite{yang2022vision}, ALBEF \cite{li2021align}, and CLIP \cite{radford2021learning}. ALBEF is composed of an image encoder, a text encoder, and a multi-modal encoder. The image encoder uses ViT-B/16 \cite{dosovitskiy2021an}, while the text and multi-modal encoders utilize a 12-layer BERT model \cite{devlin2019bert}, with the first 6 layers serving as the text encoder and the latter 6 as the multi-modal encoder. TCL, based on ALBEF, introduces three types of contrastive learning losses. CLIP consists of two main components: an image encoder and a text encoder. Both components are trained to map input images to the same feature space, bringing similar images and text closer in the feature space. CLIP offers two different image encoder options, CLIP$_{\text{ViT}}$ and CLIP$_{\text{CNN}}$, with CLIP$_{\text{ViT}}$ using ViTB/16 as the backbone of the image encoder, and CLIP$_{\text{CNN}}$ using ResNet-101 \cite{he2016deep} as the backbone of the image encoder.

\noindent \textbf{Baselines.} We use the following commonly used adversarial attack methods on VLP models as baselines: PGD \cite{madry2017towards}, BERT-Attack \cite{li2020bert}, Sep-Attack \cite{zhang2022towards}, Co-Attack \cite{zhang2022towards}, and SGA \cite{lu2023set}. The hyper-parameters for each baseline are set to their default values as provided in the open-source code.

\noindent \textbf{Metrics.} Following \cite{lu2023set}, we use the attack success rate as a metric to measure adversarial attack performance in both white-box and black-box scenarios. Specifically, for image-to-text retrieval, we use term \textbf{``TR R@K''} to denote ``the percentage of adversarial images for which the correct result is not found in the top-K (K = 1, 5, 10) texts retrieved by the VLP model $f(\cdot)$''. For text-to-image retrieval, we use term \textbf{``IR R@K''} to denote ``the percentage of adversarial texts for which the correct result is not found in the top-K (K = 1, 5, 10) images retrieved by the VLP model $f(\cdot)$''.


\noindent \textbf{Implementation details.} The hyper-parameters of our method are as follows: $A_i = A_t = 4, k = 10, \epsilon_{t} = 1, \epsilon_{x} = \frac{2}{255}, \alpha = \frac{0.5}{255}, T=10$. The reasons for setting $A_i$ and $A_x$ to these values will be explained in Section \ref{sec_ab}. The values for $\epsilon_{x}, \epsilon_{t}, \alpha$, and $T$ are set according to the default values in the open-source code from \cite{lu2023set}. The scale image method used in Algorithm \ref{alg_img} adopts the default settings of the open-source code from \cite{lu2023set}. Our method is implemented by PyTorch.

\subsection{Attack Transferability} \label{sec_exp_transfer}

To verify the effectiveness of our method, we focused on image-text retrieval and conduct comparative experiments on the Flickr30K and COCO datasets, aiming to compare our SA-Attack with baselines, i.e., PGD \cite{madry2017towards}, BERT-Attack \cite{li2020bert}, Sep-Attack \cite{zhang2022towards}, Co-Attack \cite{zhang2022towards}, and SGA \cite{lu2023set}. The experimental results are shown in Tables \ref{tab_trans_1}, \ref{tab_trans_5}, and \ref{tab_trans_10}. 

It’s not difficult to see that our SA-Attack method has shown improvements across all evaluation metrics used in the experiment, compared to the baselines. This validates the accuracy of our analysis in Section \ref{sec_motivation}. This upward trend has been verified across different datasets (FLickr30K and COCO) and different model architectures (ALBEF, TCL, CLIP$_{\text{ViT}}$ and CLIP$_{\text{CNN}}$), although there are some differences depending on the specific dataset and model selection. A detailed analysis is presented as follows.

The SA-Attack demonstrates a more prominent performance in terms of ASRs on R@1, R@5, and R@10 compared to the baselines. This indicates that SA-Attack possesses a strong attack capability.

To investigate the impact of model architectures on SA-Attack, we conducted cross-attacks on four models: ALBEF, TCL, CLIP$_{\text{ViT}}$, and CLIP$_{\text{CNN}}$. Excitingly, SA-Attack achieved performance improvements across these diverse architectures. For example, when the dataset is Flickr30K, the source model is ALBEF, and the target model is CLIP$_{\text{CNN}}$, the ASR IR R@1 for PGD, BERT-Attack, Sep-Attack, Co-Attack, SGA, SA-Attack in Table \ref{tab_trans_1} are [14.96, 46.11, 46.07, 38.89, 46.86, 49.33] respectively, and the ASR TR R@1 are [9.71, 32.69, 31.67, 25.12, 36.40, 37.93] respectively. In Table \ref{tab_trans_5}, the ASR IR R@5 for PGD, BERT-Attack, Sep-Attack, Co-Attack, SGA, SA-Attack are [5.75, 28.46, 28.21, 15.76, 29.21, 31.64] respectively, and the ASR TR R@5 are [3.07, 15.43, 14.48, 8.42, 17.02, 17.65] respectively. Similarly, when the dataset is COCO, the source model is ALBEF, and the target model is CLIP$_{\text{CNN}}$, the ASR IR R@1 for PGD, BERT-Attack, Sep-Attack, Co-Attack, SGA, SA-Attack in Table \ref{tab_trans_1} are [23.75, 65.01, 65.07, 55.64, 66.67, 68.28] respectively, and the ASR TR R@1 are [17.12, 56.11, 57.13, 47.30, 58.48, 60.24] respectively. In Table \ref{tab_trans_5}, the ASR IR R@5 for PGD, BERT-Attack, Sep-Attack, Co-Attack, SGA, SA-Attack are [12.91, 49.55, 50.24, 41.48, 52.39, 53.27] respectively, and the ASR TR R@5 are [8.95, 37.83, 38.16, 29.89, 42.21, 43.49] respectively. This suggests that, compared to the baselines, SA-Attack can adapt better to various VLP models. 

Moreover, the validation results indicate that the performance improvement of SA-Attack is somewhat universal across two different datasets, Flickr30K and COCO. This universality makes SA-Attack a potential threat, as it may exert a certain level of attack capability in various different application scenarios. 

In summary, the experimental results demonstrate the effectiveness of our method across different datasets and model architectures. We hope that our method can contribute to the security issues of future VLP models, and can help researchers gain a more comprehensive understanding of the mechanisms of adversarial transferability in VLP models.

\section{Discussion}

\subsection{Cross-task Transferability}

To further discuss the effectiveness of our method, we use the task of Visual Grounding (VG), which is to identify specific objects in an image, as an example to explore the cross-task transferability of our method. Specifically, we input an image and a description of the object (such as a sentence, title, or description) into the VLP model, and the VLP model outputs a box that marks the object mentioned in the description. This may sound similar to object detection, but VG incorporates language information into the input, which object detection does not. 

Table \ref{tab_vg} presents the experimental results, where adversarial image, which are crafted from the Image-Text Retrieval (ITR), are used to attack the VG on RefCOCO+ dataset \cite{yu2016modeling,kazemzadeh2014refer}. Both the source model and the target model are ALBEF. The baseline refers to the initial VG performance on clean data. It is not difficult to see from Table \ref{tab_vg} that the values of our method are lower than those of the baseline, indicating that our method has a certain cross-task transferability. This also reveals that, similar to the face recognition \cite{li2023sibling}, there is a certain correlation between different VLP tasks. In other words, an adversarial example in one task may affect the performance of other tasks.

\subsection{Ablation Studies} \label{sec_ab}

\noindent  \textbf{Ablation Studies on $A_x$.} Table \ref{tab_ab_ax} reports the results of ablation studies on $A_x$, with the value of $A_t$ fixed at $4$. The table presents the attack success rates (R@1) on the Flickr30K and COCO datasets. The gray background in the table indicates white-box attacks. 

\noindent  \textbf{Ablation Studies on $A_t$.} Table \ref{tab_ab_at} reports the results of ablation studies on $A_t$, with the value of $A_x$ fixed at $4$. The table presents the attack success rates (R@1) on the Flickr30K and COCO datasets. The gray background in the table indicates white-box attacks. The results from Table \ref{tab_ab_ax} and Table \ref{tab_ab_at} indicate that the differences in outcomes for various $A_x$ and $A_t$ are minimal, with fluctuations of approximately 2\%. For example, in Table \ref{tab_ab_ax}, when the dataset is the Flickr30K, and the source model is TCL while the target model is CLIP$_{\text{CNN}}$, the ASRs for TR R@1 are $39.21 (A_x=4), 40.61 (A_x=8),$ and $40.10 (A_x=12)$ respectively. Similarly, the ASRs for IR R@1 are $50.77 (A_x=4), 50.12 (A_x=8),$ and $50.67 (A_x=12)$ respectively. These results suggest that our method has the following advantages: \ding{182} Our method does not require spending a lot of time to select the optimal hyper-parameters; \ding{183} As shown in Table \ref{tab_time}, in our experiments conducted on the NVIDIA GeForce RTX 4090 GPU, the runtime increases with the increase of both $A_t$ and $A_x$. Therefore, for convenience, we set the hyper-parameters $A_x$ and $A_t$ to 4 in Section \ref{sec_setup}.

\subsection{Visualization}

Figures \ref{fig_vis_flickr} and Figrue \ref{fig_vis_coco} respectively showcase the visualization results of our method on the Flickr30K and COCO datasets. In the listed adversarial text, the modified words are displayed in red bold font. The performance appears to be impressive. Specifically, in terms of images, the perturbations generated by our method are minimal, almost imperceptible to the naked eye. As for texts, our method only alters a few words. 


\section{Conclusion}

In this paper, we take input diversity as a starting point and devise a transfer attack method for VLP models, proposing a self-augmentation attack method (dubbed SA-Attack). In essence, we incorporate augmented input images and input texts separately when generating adversarial examples for image and text modalities. Through a series of experiments, we validate the effectiveness of this method. We hope that our SA-Attack will provide new perspectives and insights for designing transferable adversarial attack methods on VLP models. In future work, we plan to deepen our understanding of SA-Attack and explore its applicability in a broader range of tasks and application scenarios. 

\bibliography{egbib}{} 

\begin{thebibliography}{10}
\providecommand{\url}[1]{#1}
\csname url@samestyle\endcsname
\providecommand{\newblock}{\relax}
\providecommand{\bibinfo}[2]{#2}
\providecommand{\BIBentrySTDinterwordspacing}{\spaceskip=0pt\relax}
\providecommand{\BIBentryALTinterwordstretchfactor}{4}
\providecommand{\BIBentryALTinterwordspacing}{\spaceskip=\fontdimen2\font plus
\BIBentryALTinterwordstretchfactor\fontdimen3\font minus \fontdimen4\font\relax}
\providecommand{\BIBforeignlanguage}[2]{{%
\expandafter\ifx\csname l@#1\endcsname\relax
\typeout{** WARNING: IEEEtran.bst: No hyphenation pattern has been}%
\typeout{** loaded for the language `#1'. Using the pattern for}%
\typeout{** the default language instead.}%
\else
\language=\csname l@#1\endcsname
\fi
#2}}
\providecommand{\BIBdecl}{\relax}
\BIBdecl

\bibitem{yin2023vlattack}
Z.~Yin, M.~Ye, T.~Zhang, T.~Du, J.~Zhu, H.~Liu, J.~Chen, T.~Wang, and F.~Ma, ``Vlattack: Multimodal adversarial attacks on vision-language tasks via pre-trained models,'' \emph{NeurIPS}, 2023.

\bibitem{zhou2023advclip}
Z.~Zhou, S.~Hu, M.~Li, H.~Zhang, Y.~Zhang, and H.~Jin, ``Advclip: Downstream-agnostic adversarial examples in multimodal contrastive learning,'' in \emph{ACM MM'23}, 2023.

\bibitem{zhang2022towards}
J.~Zhang, Q.~Yi, and J.~Sang, ``Towards adversarial attack on vision-language pre-training models,'' in \emph{Proceedings of the 30th ACM International Conference on Multimedia}, 2022.

\bibitem{cao2023stylefool}
Y.~Cao, X.~Xiao, R.~Sun, D.~Wang, M.~Xue, and S.~Wen, ``Stylefool: Fooling video classification systems via style transfer,'' in \emph{SP}, 2023.

\bibitem{hingun2023reap}
N.~Hingun, C.~Sitawarin, J.~Li, and D.~Wagner, ``Reap: A large-scale realistic adversarial patch benchmark,'' in \emph{ICCV}, 2023.

\bibitem{jiang2023efficient}
K.~Jiang, Z.~Chen, H.~Huang, J.~Wang, D.~Yang, B.~Li, Y.~Wang, and W.~Zhang, ``Efficient decision-based black-box patch attacks on video recognition,'' in \emph{ICCV}, 2023.

\bibitem{bao2022vlmo}
H.~Bao, W.~Wang, L.~Dong, Q.~Liu, O.~K. Mohammed, K.~Aggarwal, S.~Som, S.~Piao, and F.~Wei, ``Vlmo: Unified vision-language pre-training with mixture-of-modality-experts,'' \emph{NeurIPS}, 2022.

\bibitem{wang2023position}
J.~Wang, P.~Zhou, M.~Z. Shou, and S.~Yan, ``Position-guided text prompt for vision-language pre-training,'' in \emph{CVPR}, 2023.

\bibitem{wu2023revisiting}
W.~Wu, Z.~Sun, and W.~Ouyang, ``Revisiting classifier: Transferring vision-language models for video recognition,'' in \emph{AAAI}, 2023.

\bibitem{xue2023diffusion}
H.~Xue, A.~Araujo, B.~Hu, and Y.~Chen, ``Diffusion-based adversarial sample generation for improved stealthiness and controllability,'' \emph{arXiv preprint arXiv:2305.16494}, 2023.

\bibitem{qiu2020semanticadv}
H.~Qiu, C.~Xiao, L.~Yang, X.~Yan, H.~Lee, and B.~Li, ``Semanticadv: Generating adversarial examples via attribute-conditioned image editing,'' in \emph{ECCV}, 2020, pp. 19--37.

\bibitem{morris220textattack}
J.~X. Morris, E.~Lifland, J.~Y. Yoo, J.~Grigsby, D.~Jin, and Y.~Qi, ``Textattack: {A} framework for adversarial attacks, data augmentation, and adversarial training in {NLP},'' in \emph{EMNLP}, 2020.

\bibitem{li2020bert}
L.~Li, R.~Ma, Q.~Guo, X.~Xue, and X.~Qiu, ``{BERT-ATTACK:} adversarial attack against {BERT} using {BERT},'' in \emph{EMNLP}, 2020.

\bibitem{jin2020bert}
D.~Jin, Z.~Jin, J.~T. Zhou, and P.~Szolovits, ``Is bert really robust? a strong baseline for natural language attack on text classification and entailment,'' in \emph{AAAI}, 2020.

\bibitem{wei2023towards}
Z.~Wei, J.~Chen, M.~Goldblum, Z.~Wu, T.~Goldstein, Y.-G. Jiang, and L.~S. Davis, ``Towards transferable adversarial attacks on image and video transformers.'' \emph{TIP}, 2023.

\bibitem{chen2023gcma}
K.~Chen, Z.~Wei, J.~Chen, Z.~Wu, and Y.-G. Jiang, ``Gcma: Generative cross-modal transferable adversarial attacks from images to videos,'' in \emph{ACM MM}, 2023.

\bibitem{zhang2023transferable}
J.~Zhang, Y.~Huang, W.~Wu, and M.~R. Lyu, ``Transferable adversarial attacks on vision transformers with token gradient regularization,'' in \emph{CVPR}, 2023.

\bibitem{lu2023set}
D.~Lu, Z.~Wang, T.~Wang, W.~Guan, H.~Gao, and F.~Zheng, ``Set-level guidance attack: Boosting adversarial transferability of vision-language pre-training models,'' in \emph{ICCV}, 2023.

\bibitem{zhu2022toward}
Y.~Zhu, Y.~Chen, X.~Li, K.~Chen, Y.~He, X.~Tian, B.~Zheng, Y.~Chen, and Q.~Huang, ``Toward understanding and boosting adversarial transferability from a distribution perspective,'' \emph{TIP}, 2022.

\bibitem{wang2023structure}
X.~Wang, Z.~Zhang, and J.~Zhang, ``Structure invariant transformation for better adversarial transferability,'' in \emph{ICCV}, 2023.

\bibitem{byun2022improving}
J.~Byun, S.~Cho, M.-J. Kwon, H.-S. Kim, and C.~Kim, ``Improving the transferability of targeted adversarial examples through object-based diverse input,'' in \emph{CVPR}, 2022.

\bibitem{wang2021admix}
X.~Wang, X.~He, J.~Wang, and K.~He, ``Admix: Enhancing the transferability of adversarial attacks,'' in \emph{ICCV}, 2021.

\bibitem{wei2019eda}
J.~W. Wei and K.~Zou, ``{EDA:} easy data augmentation techniques for boosting performance on text classification tasks,'' in \emph{EMNLP-IJCNLP}, 2019.

\bibitem{madry2017towards}
A.~Madry, A.~Makelov, L.~Schmidt, D.~Tsipras, and A.~Vladu, ``Towards deep learning models resistant to adversarial attacks,'' in \emph{ICLR}, 2018.

\bibitem{plummer2015flickr30k}
B.~A. Plummer, L.~Wang, C.~M. Cervantes, J.~C. Caicedo, J.~Hockenmaier, and S.~Lazebnik, ``Flickr30k entities: Collecting region-to-phrase correspondences for richer image-to-sentence models,'' in \emph{ICCV}, 2015.

\bibitem{lin2014microsoft}
T.-Y. Lin, M.~Maire, S.~Belongie, J.~Hays, P.~Perona, D.~Ramanan, P.~Doll{\'a}r, and C.~L. Zitnick, ``Microsoft coco: Common objects in context,'' in \emph{ECCV}, 2014.

\bibitem{yang2022vision}
J.~Yang, J.~Duan, S.~Tran, Y.~Xu, S.~Chanda, L.~Chen, B.~Zeng, T.~Chilimbi, and J.~Huang, ``Vision-language pre-training with triple contrastive learning,'' in \emph{CVPR}, 2022.

\bibitem{li2021align}
J.~Li, R.~Selvaraju, A.~Gotmare, S.~Joty, C.~Xiong, and S.~C.~H. Hoi, ``Align before fuse: Vision and language representation learning with momentum distillation,'' \emph{NeurIPS}, 2021.

\bibitem{radford2021learning}
A.~Radford, J.~W. Kim, C.~Hallacy, A.~Ramesh, G.~Goh, S.~Agarwal, G.~Sastry, A.~Askell, P.~Mishkin, J.~Clark \emph{et~al.}, ``Learning transferable visual models from natural language supervision,'' in \emph{ICML}, 2021.

\bibitem{chen2023vlp}
F.-L. Chen, D.-Z. Zhang, M.-L. Han, X.-Y. Chen, J.~Shi, S.~Xu, and B.~Xu, ``Vlp: A survey on vision-language pre-training,'' \emph{MIR}, 2023.

\bibitem{huang2021seeing}
Z.~Huang, Z.~Zeng, Y.~Huang, B.~Liu, D.~Fu, and J.~Fu, ``Seeing out of the box: End-to-end pre-training for vision-language representation learning,'' in \emph{CVPR}, 2021.

\bibitem{ji2023seeing}
Y.~Ji, R.~Tu, J.~Jiang, W.~Kong, C.~Cai, W.~Zhao, H.~Wang, Y.~Yang, and W.~Liu, ``Seeing what you miss: Vision-language pre-training with semantic completion learning,'' in \emph{CVPR}, 2023.

\bibitem{ji2023map}
Y.~Ji, J.~Wang, Y.~Gong, L.~Zhang, Y.~Zhu, H.~Wang, J.~Zhang, T.~Sakai, and Y.~Yang, ``Map: Multimodal uncertainty-aware vision-language pre-training model,'' in \emph{CVPR}, 2023.

\bibitem{kurakin2016adversarial}
A.~Kurakin, I.~Goodfellow, and S.~Bengio, ``Adversarial machine learning at scale,'' in \emph{ICLR}, 2017.

\bibitem{gu2023survey}
J.~Gu, X.~Jia, P.~de~Jorge, W.~Yu, X.~Liu, A.~Ma, Y.~Xun, A.~Hu, A.~Khakzar, Z.~Li \emph{et~al.}, ``A survey on transferability of adversarial examples across deep neural networks,'' \emph{arXiv preprint arXiv:2310.17626}, 2023.

\bibitem{yu2023reliable}
W.~Yu, J.~Gu, Z.~Li, and P.~Torr, ``Reliable evaluation of adversarial transferability,'' \emph{arXiv preprint arXiv:2306.08565}, 2023.

\bibitem{xie2019improving}
C.~Xie, Z.~Zhang, Y.~Zhou, S.~Bai, J.~Wang, Z.~Ren, and A.~L. Yuille, ``Improving transferability of adversarial examples with input diversity,'' in \emph{CVPR}, 2019.

\bibitem{dong2019evading}
Y.~Dong, T.~Pang, H.~Su, and J.~Zhu, ``Evading defenses to transferable adversarial examples by translation-invariant attacks,'' in \emph{CVPR}, 2019.

\bibitem{zou2020improving}
J.~Zou, Z.~Pan, J.~Qiu, X.~Liu, T.~Rui, and W.~Li, ``Improving the transferability of adversarial examples with resized-diverse-inputs, diversity-ensemble and region fitting,'' in \emph{ECCV}, 2020.

\bibitem{lin2020nesterov}
J.~Lin, C.~Song, K.~He, L.~Wang, and J.~E. Hopcroft, ``Nesterov accelerated gradient and scale invariance for adversarial attacks,'' in \emph{ICLR}, 2020.

\bibitem{huang2022transferable}
Y.~Huang and A.~W. Kong, ``Transferable adversarial attack based on integrated gradients,'' in \emph{ICLR}, 2022.

\bibitem{sundararajan2017axiomatic}
M.~Sundararajan, A.~Taly, and Q.~Yan, ``Axiomatic attribution for deep networks,'' in \emph{ICML}, 2017.

\bibitem{dong2018boosting}
Y.~Dong, F.~Liao, T.~Pang, H.~Su, J.~Zhu, X.~Hu, and J.~Li, ``Boosting adversarial attacks with momentum,'' in \emph{CVPR}, 2018.

\bibitem{nesterov1983method}
Y.~Nesterov, ``A method for unconstrained convex minimization problem with the rate of convergence o (1/k2),'' in \emph{Dokl. Akad. Nauk. SSSR}, 1983.

\bibitem{zou2022making}
J.~Zou, Y.~Duan, B.~Li, W.~Zhang, Y.~Pan, and Z.~Pan, ``Making adversarial examples more transferable and indistinguishable,'' in \emph{AAAI}, 2022.

\bibitem{kingma20156adam}
D.~P. Kingma and J.~Ba, ``Adam: A method for stochastic optimization,'' in \emph{ICLR}, 2015.

\bibitem{wang2021enhancing}
X.~Wang and K.~He, ``Enhancing the transferability of adversarial attacks through variance tuning,'' in \emph{CVPR}, 2021.

\bibitem{qin2022boosting}
Z.~Qin, Y.~Fan, Y.~Liu, L.~Shen, Y.~Zhang, J.~Wang, and B.~Wu, ``Boosting the transferability of adversarial attacks with reverse adversarial perturbation,'' \emph{NeurIPS}, 2022.

\bibitem{zhu2023boosting}
H.~Zhu, Y.~Ren, X.~Sui, L.~Yang, and W.~Jiang, ``Boosting adversarial transferability via gradient relevance attack,'' in \emph{ICCV}, 2023.

\bibitem{zhou2018transferable}
W.~Zhou, X.~Hou, Y.~Chen, M.~Tang, X.~Huang, X.~Gan, and Y.~Yang, ``Transferable adversarial perturbations,'' in \emph{ECCV}, 2018.

\bibitem{waseda2023closer}
F.~Waseda, S.~Nishikawa, T.-N. Le, H.~H. Nguyen, and I.~Echizen, ``Closer look at the transferability of adversarial examples: How they fool different models differently,'' in \emph{WACV}, 2023.

\bibitem{li2023improving}
Q.~Li, Y.~Guo, W.~Zuo, and H.~Chen, ``Improving adversarial transferability by intermediate-level perturbation decay,'' \emph{NeurIPS}, 2023.

\bibitem{zhang2022improving}
J.~Zhang, W.~Wu, J.-t. Huang, Y.~Huang, W.~Wang, Y.~Su, and M.~R. Lyu, ``Improving adversarial transferability via neuron attribution-based attacks,'' in \emph{CVPR}, 2022.

\bibitem{wu2020skip}
D.~Wu, Y.~Wang, S.~Xia, J.~Bailey, and X.~Ma, ``Skip connections matter: On the transferability of adversarial examples generated with resnets,'' in \emph{ICLR}, 2020.

\bibitem{benz2021batch}
P.~Benz, C.~Zhang, and I.~S. Kweon, ``Batch normalization increases adversarial vulnerability and decreases adversarial transferability: A non-robust feature perspective,'' in \emph{ICCV}, 2021.

\bibitem{naseer2022on}
M.~Naseer, K.~Ranasinghe, S.~Khan, F.~S. Khan, and F.~Porikli, ``On improving adversarial transferability of vision transformers,'' in \emph{ICLR}, 2022.

\bibitem{huang2023t}
H.~Huang, Z.~Chen, H.~Chen, Y.~Wang, and K.~Zhang, ``T-sea: Transfer-based self-ensemble attack on object detection,'' in \emph{CVPR}, 2023.

\bibitem{li2020towards}
M.~Li, C.~Deng, T.~Li, J.~Yan, X.~Gao, and H.~Huang, ``Towards transferable targeted attack,'' in \emph{CVPR}, 2020.

\bibitem{xiao2021improving}
Z.~Xiao, X.~Gao, C.~Fu, Y.~Dong, W.~Gao, X.~Zhang, J.~Zhou, and J.~Zhu, ``Improving transferability of adversarial patches on face recognition with generative models,'' in \emph{CVPR}, 2021.

\bibitem{zhang2022investigating}
C.~Zhang, P.~Benz, A.~Karjauv, J.~W. Cho, K.~Zhang, and I.~S. Kweon, ``Investigating top-k white-box and transferable black-box attack,'' in \emph{CVPR}, 2022.

\bibitem{fang2022learning}
S.~Fang, J.~Li, X.~Lin, and R.~Ji, ``Learning to learn transferable attack,'' in \emph{AAAI}, 2022.

\bibitem{li2023making}
Q.~Li, Y.~Guo, W.~Zuo, and H.~Chen, ``Making substitute models more bayesian can enhance transferability of adversarial examples,'' in \emph{ICLR}, 2023.

\bibitem{ma2023transferable}
W.~Ma, Y.~Li, X.~Jia, and W.~Xu, ``Transferable adversarial attack for both vision transformers and convolutional networks via momentum integrated gradients,'' in \emph{ICCV}, 2023.

\bibitem{wei2022towards}
Z.~Wei, J.~Chen, M.~Goldblum, Z.~Wu, T.~Goldstein, and Y.-G. Jiang, ``Towards transferable adversarial attacks on vision transformers,'' in \emph{AAAI}, 2022.

\bibitem{mahmood2021robustness}
K.~Mahmood, R.~Mahmood, and M.~Van~Dijk, ``On the robustness of vision transformers to adversarial examples,'' in \emph{ICCV}, 2021.

\bibitem{li2023sibling}
Z.~Li, B.~Yin, T.~Yao, J.~Guo, S.~Ding, S.~Chen, and C.~Liu, ``Sibling-attack: Rethinking transferable adversarial attacks against face recognition,'' in \emph{CVPR}, 2023.

\bibitem{jia2022adv}
S.~Jia, B.~Yin, T.~Yao, S.~Ding, C.~Shen, X.~Yang, and C.~Ma, ``Adv-attribute: Inconspicuous and transferable adversarial attack on face recognition,'' \emph{NeurIPS}, 2022.

\bibitem{yin2021adv}
B.~Yin, W.~Wang, T.~Yao, J.~Guo, Z.~Kong, S.~Ding, J.~Li, and C.~Liu, ``Adv-makeup: A new imperceptible and transferable attack on face recognition,'' \emph{IJCAI}, 2021.

\bibitem{wang2023exploring}
Y.~Wang, W.~Hu, Y.~Dong, and R.~Hong, ``Exploring transferability of multimodal adversarial samples for vision-language pre-training models with contrastive learning,'' \emph{arXiv preprint arXiv:2308.12636}, 2023.

\bibitem{zhu2023ai}
H.~Zhu, S.~Zhang, and K.~Chen, ``Ai-guardian: Defeating adversarial attacks using backdoors,'' in \emph{SP}, 2023.

\bibitem{dyrmishi2023empirical}
S.~Dyrmishi, S.~Ghamizi, T.~Simonetto, Y.~Le~Traon, and M.~Cordy, ``On the empirical effectiveness of unrealistic adversarial hardening against realistic adversarial attacks,'' in \emph{SP}, 2023.

\bibitem{zhang2022adversarial}
R.~Zhang, J.~Liu, Y.~Ding, Z.~Wang, Q.~Wu, and K.~Ren, ``“adversarial examples” for proof-of-learning,'' in \emph{SP}, 2022.

\bibitem{liu2023prompt}
Y.~Liu, G.~Deng, Y.~Li, K.~Wang, T.~Zhang, Y.~Liu, H.~Wang, Y.~Zheng, and Y.~Liu, ``Prompt injection attack against llm-integrated applications,'' \emph{arXiv preprint arXiv:2306.05499}, 2023.

\bibitem{deng2023jailbreaker}
G.~Deng, Y.~Liu, Y.~Li, K.~Wang, Y.~Zhang, Z.~Li, H.~Wang, T.~Zhang, and Y.~Liu, ``Jailbreaker: Automated jailbreak across multiple large language model chatbots,'' \emph{arXiv preprint arXiv:2307.08715}, 2023.

\bibitem{si2022so}
W.~M. Si, M.~Backes, J.~Blackburn, E.~De~Cristofaro, G.~Stringhini, S.~Zannettou, and Y.~Zhang, ``Why so toxic? measuring and triggering toxic behavior in open-domain chatbots,'' in \emph{CCS}, 2022.

\bibitem{zhang2017mixup}
H.~Zhang, M.~Ciss{\'{e}}, Y.~N. Dauphin, and D.~Lopez{-}Paz, ``mixup: Beyond empirical risk minimization,'' in \emph{ICLR}, 2018.

\bibitem{yun2019cutmix}
S.~Yun, D.~Han, S.~J. Oh, S.~Chun, J.~Choe, and Y.~Yoo, ``Cutmix: Regularization strategy to train strong classifiers with localizable features,'' in \emph{ICCV}, 2019.

\bibitem{cubuk2018autoaugment}
E.~D. Cubuk, B.~Zoph, D.~Mane, V.~Vasudevan, and Q.~V. Le, ``Autoaugment: Learning augmentation policies from data,'' \emph{arXiv preprint arXiv:1805.09501}, 2018.

\bibitem{ho2019population}
D.~Ho, E.~Liang, X.~Chen, I.~Stoica, and P.~Abbeel, ``Population based augmentation: Efficient learning of augmentation policy schedules,'' in \emph{ICML}, 2019.

\bibitem{lingchen2020uniformaugment}
T.~C. LingChen, A.~Khonsari, A.~Lashkari, M.~R. Nazari, J.~S. Sambee, and M.~A. Nascimento, ``Uniformaugment: A search-free probabilistic data augmentation approach,'' \emph{arXiv preprint arXiv:2003.14348}, 2020.

\bibitem{muller2021trivialaugment}
S.~G. M{\"u}ller and F.~Hutter, ``Trivialaugment: Tuning-free yet state-of-the-art data augmentation,'' in \emph{ICCV}, 2021.

\bibitem{ban2022pre}
Y.~Ban and Y.~Dong, ``Pre-trained adversarial perturbations,'' in \emph{NeurIPS}, 2022.

\bibitem{dosovitskiy2021an}
A.~Dosovitskiy, L.~Beyer, A.~Kolesnikov, D.~Weissenborn, X.~Zhai, T.~Unterthiner, M.~Dehghani, M.~Minderer, G.~Heigold, S.~Gelly, J.~Uszkoreit, and N.~Houlsby, ``An image is worth 16x16 words: Transformers for image recognition at scale,'' in \emph{ICLR}, 2021.

\bibitem{devlin2019bert}
J.~Devlin, M.~Chang, K.~Lee, and K.~Toutanova, ``{BERT:} pre-training of deep bidirectional transformers for language understanding,'' in \emph{NAACL-HLT}.\hskip 1em plus 0.5em minus 0.4em\relax Association for Computational Linguistics, 2019.

\bibitem{he2016deep}
K.~He, X.~Zhang, S.~Ren, and J.~Sun, ``Deep residual learning for image recognition,'' in \emph{CVPR}, 2016.

\bibitem{yu2016modeling}
L.~Yu, P.~Poirson, S.~Yang, A.~C. Berg, and T.~L. Berg, ``Modeling context in referring expressions,'' in \emph{ECCV}, B.~Leibe, J.~Matas, N.~Sebe, and M.~Welling, Eds., 2016.

\bibitem{kazemzadeh2014refer}
S.~Kazemzadeh, V.~Ordonez, M.~Matten, and T.~L. Berg, ``Referitgame: Referring to objects in photographs of natural scenes,'' in \emph{EMNLP}, A.~Moschitti, B.~Pang, and W.~Daelemans, Eds., 2014.

\end{thebibliography}
\bibliographystyle{IEEEtran}




\end{document}